# Dynamic Gait Modelling of Lower Limb Dynamics A Mathematical Approach

Barath Kumar JK and Aswadh Khumar G S

*Abstract*— This paper focuses on the analysis of human gait cycle dynamics and presents a mathematical model to determine the torque exerted on the lower limb joints throughout the complete gait cycle, including its various phases. The study involved a healthy subject who participated in a series of initial walking experiments. The development of a mathematical model that accurately represents the natural motion of the human lower limb has garnered significant attention in the field of lower limb prosthetics design. In this study, the researchers incorporated the functional relationship between the limb joints and the end-effector of the lower extremity. This knowledge is crucial for rehabilitation purposes as it helps in understanding the connectivity of joints, links, and the overall body orientation required to effectively control the motion of the actuators. When analysing physical activities, measurements of human strength play a crucial role. Traditionally, these measurements have focused on determining the maximum voluntary torque at a single joint angle and angular velocity. However, it is important to consider that the available strength varies significantly with joint position and velocity. Therefore, when examining dynamic activities, strength measurements should account for these variations. To address this, the researchers present a model that represents the maximum voluntary joint torque as a function of joint angle and angular velocity. This model offers an efficient method to incorporate variations in strength with joint angle and angular velocity, allowing for more accurate comparisons between joint torques calculated using inverse dynamics and the maximum available joint torques. Based on this model, the researchers estimate various gait parameters, including the medio-lateral rotation of the lower limbs during stance and swing, stride length, and velocity. These estimations are achieved through the integration of angular velocity data.

*Index Terms*— IK, kinematics, dynamics, joint angle estimation

## I. INTRODUCTION

The mathematical model, which uses the prosthesis's own mechanical principles, is the alternative strategy. In a study referenced as [5], researchers established a neuromuscular model of human motion and designed a prosthesis controller that incorporates muscle reflexes and local feedback. Through simulations, they found that the neuromuscular model control resulted in a more robust gait compared to the impedance control method, and it also improved balance recovery in individuals with amputations. However, neuromuscular models are not suitable for transfemoral prostheses because they require considering coordinated movement between muscles, which is challenging due to the significant muscle deficits in transfemoral amputees. On the other hand, the mathematical model of the prosthesis based on Lagrange equations is commonly used. In another study mentioned as [6], the researchers derived the motion equations for an Active Knee Prosthesis using the Euler-Lagrange method.

When it comes to lower limb prosthetics, transtibial amputations [35] usually do not require actuation systems as much as transfemoral amputations [36]. Therefore, the focus of prosthetic development primarily lies in developing actuation systems for the knee mechanism in transfemoral prosthetics. The knee joint is the crucial joint that demands more attention and detailed work in prosthetic development, aiming to enable amputees to effectively carry out their daily activities and routines. In an active prosthetic knee, a high-level controller [6] is used to estimate desired knee joint positions. The user's intentions are ascertained by this high-level controller, which then transforms them into a set-point input for the low-level controller, a proportional-derivative (PD) controller. The low-level controller uses this information, along with the prosthetic joint's actual motion and other unique characteristics (such inertia and motor-torque restrictions), to determine the trajectory and actuation required to fulfil the high-level controller's directive. The types of activities being engaged in (such as walking or running), the rate of motion, and the particular gait phase (such as heel strike, midstance, and toe-off) all affect how the knee joint is positioned.

Our study focuses on investigating various innovative mathematical methods to determine the durations of the swing and stance phases within a gait cycle. Additionally, we aim to establish a relationship between the angles of the shank and knee within a restricted domain. To compute the knee torque, we employ a technique for estimating knee angles and validate the computed torque through the utilization of MATLAB Simulink, Lagrangian-Euler equations, and an open-source dataset. In order to achieve these objectives, we utilize mathematical approaches that offer new insights into analysing gait cycles. By employing these approaches, we aim to accurately determine the durations of the swing and stance phases, which are crucial components of the overall gait pattern.

Furthermore, we seek to establish a relationship between the angles of the shank and knee, focusing on a specific domain. This restricted domain allows us to derive a more precise and applicable relationship between these two parameters. To compute the knee torque, we employ a knee angle estimation technique. This technique provides us with the necessary data to calculate the torque exerted on the knee joint during the gait cycle. We use a variety of computational tools, including MATLAB Simulink, which enables us to simulate and analyse the behaviour of the system, to test the correctness and dependability of our estimated torque values. In addition, we simulate the motion and dynamics of the knee joint using Lagrangian-Euler equations, a mathematical framework widely utilized in dynamics and mechanics. We can fully comprehend the forces and torques acting on the knee joint during the gait cycle by applying these equations. We use an open-source dataset to evaluate our computed torque values, ensuring the validity and application of our findings. This dataset contains real-world measurements and observations, enabling us to compare our results with established and reliable data.

## II. DYNAMIC MODELLING OF LOWER LIMB PROSTHESIS

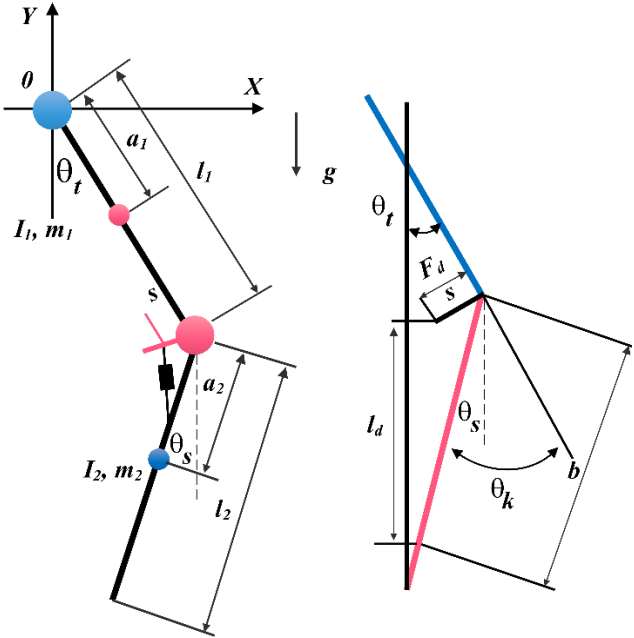

**Figure 1.** Dynamic Model of Lower Limb Prosthesis

The diagram presented in Figure 1 illustrates the dynamic model that incorporates a hydraulic damper into the lower limb prosthetic system. This system can be simplified into a two rigid body model representing the movement of the thigh and calf in the sagittal plane. The foot and leg tube are considered as a rigid connection within this model. In the diagram, the subscripts 1 and 2 correspond to the thigh and calf parameters, respectively.

| Parameters | Specifications |
|---|---|
| $m_1$ | Thigh mass |
| $m_2$ | Shank mass |
| $a_1$ | Distance of the thigh mass centre from the hip joint |
| $a_2$ | Distance of the shank mass centre from the knee joint |
| $I_1$ | Thigh moment of inertia |
| $I_2$ | Shank moment of inertia |
| $l_1$ | Thigh length |
| $l_2$ | Shank length |
| $s$ | Offset between the knee centre and location of attachment of damper piston on the thigh |
| $b$ | Distance between the knee centre and location of the damper attachment on the shank |
| $\theta_s$ | Absolute angle of shank from vertical |
| $\theta_k$ | Knee angle |
| $\theta_t$ | Absolute angle of thigh from horizontal |
| $l_d$ | Length of Damper |
| $T_1$ | Hip Torque |

**Table 1.** Parameters of the Dynamic Model

The dynamic model described in the previous statement assumes the absence of friction between the joints. To mathematically represent the dynamics of the system, the researchers (Xiaodong Wang et al., 2015) employed the second type of Lagrange equation:

$$D(\theta)\ddot{\theta} + C(\theta,\dot{\theta})\dot{\theta} + G(\theta) = \Gamma \quad (1.1)$$

The inertial matrix $D(\theta)$ can be mathematically denoted as:

$$D(\theta) = \begin{bmatrix} m_1 a_1^2 + I_1 + m_2 l_1^2 & -m_2 l_1 a_2 \cos(\theta_t + \theta_s) \\ -m_2 l_1 a_2 \cos(\theta_t + \theta_s) & m_2 a_2^2 + I_2 \end{bmatrix}$$

The Coriolis and centrifugal terms $C(\theta,\dot{\theta})$ can be denoted as:

$$C(\theta,\dot{\theta}) = \begin{bmatrix} m_2 l_1 a_2 (\dot{\theta}_s)^2 \sin(\theta_t + \theta_s) \\ m_2 l_1 a_2 (\dot{\theta}_t)^2 \sin(\theta_t + \theta_s) \end{bmatrix} \quad (1.2)$$

The gravity vector $G(\theta)$ can be expressed as:

$$G(\theta) = \begin{bmatrix} m_1 g a_1 \sin(\theta_t) + m_2 g l_1 \sin(\theta_t) \\ m_2 g a_2 \sin(\theta_s) \end{bmatrix} \quad (1.3)$$

The hip and knee input parameter matrices is given by:

$$\Gamma = \begin{bmatrix} T_1 + F_d b \sin(\theta_s - \beta) \\ -F_d b \sin(\theta_s - \beta) \end{bmatrix} \quad (1.4)$$

The thigh and calf angle vector matrix $\theta$ can be expressed as:

$$\theta = \begin{bmatrix} \theta_t \\ \theta_s \end{bmatrix} \quad (1.5)$$

The Knee torque generated during swing phase can be calculated by:

$$\tau_k = m_2 a_2^2 \ddot{\theta}_s - m_2 l_1 a_2 \ddot{\theta}_t \cos(\theta_t + \theta_s) + m_2 l_1 a_2 (\dot{\theta}_t)^2 \sin(\theta_t + \theta_s) \\ + I_2 \ddot{\theta}_s + m_2 g a_2 \sin(\theta_s) \quad (1.6)$$

### III. DOMAIN CONSTRAINED JOINT ANGLE RELATIONSHIP

Establishing a domain-restricted relationship between $\theta_{shank}$ and $\theta_k$,

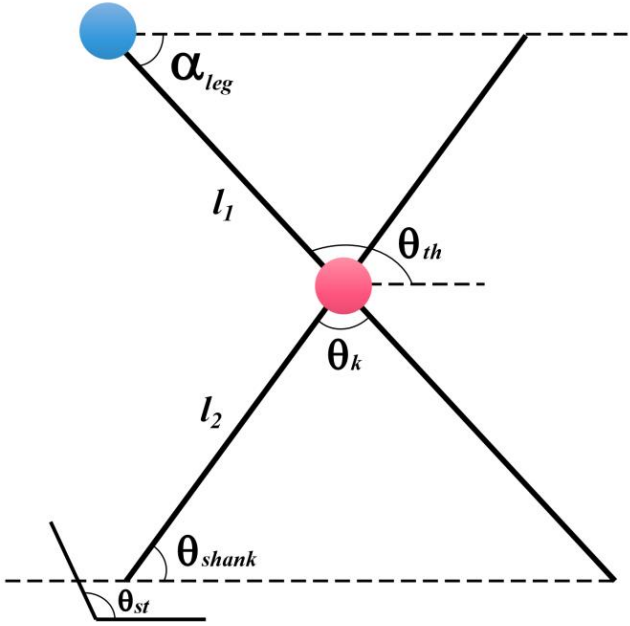

**Figure 2.** Geometric Model of Lower Limb Prosthesis

By Alternate interior angles:

$$\theta_x = 180° - \theta_{th} \quad (2.1)$$

$$\alpha = 180° - \theta_x - \theta_{lg} \quad (2.2)$$

$$\alpha = \theta_{th} - \theta_{lg} = \theta_k \quad (2.3)$$

Here $\theta_{leg}$ is same as $\theta_{shank}$,

$\alpha_{leg}$ is defined as the angle between the hip line (Horizontal) and the line connecting hip joint and the ankle joint, now assume an instance when $\alpha_{leg}$ and $\theta_{shank}$ lie in the same horizontal,

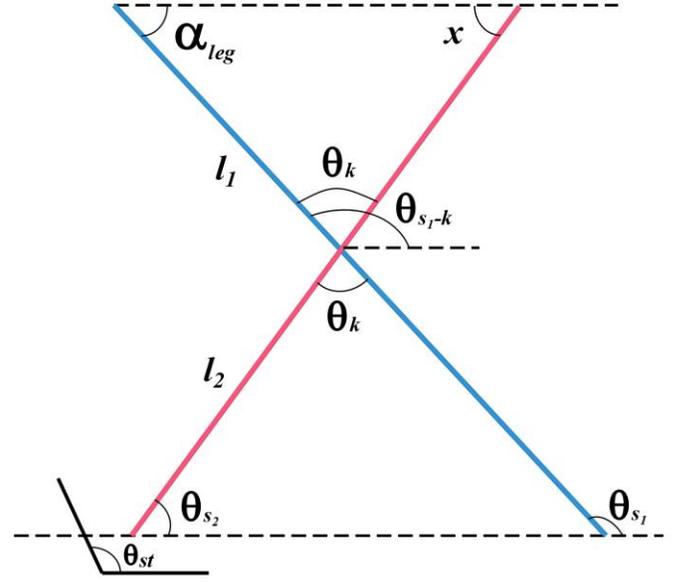

**Figure 3.** Same Horizontal Condition

$x = \theta s_1$, (Alternate interior angles)

By triangle sum property:

$$\alpha + \theta_k + \theta s_1 - \theta_k = 180° \quad (2.4)$$

$\alpha_{leg} = 180° - \theta s_1$; if they lie in the same horizontal leg,

$$\theta s_2 + 180° - \theta s_1 + \theta_k = 180° \quad (2.5)$$
$$\theta s_1 = (\theta_k + \theta s_2) \quad (2.6)$$
$$\alpha_{leg} = 180° - (\theta_k + \theta s_2) \quad (2.7)$$

Where $\theta s_2$ is any angle; other than $180° - \theta s_1 = \alpha_{leg}$

$$\alpha_{leg} = \cos^{-1}\left(\frac{\sin \theta_k}{l_s l}\right) - \theta_k \quad (2.8)$$

from (1) and (2);

$$180 - \theta_k - \theta_{s_2} = \cos^{-1}\left(\frac{\sin \theta_k}{l_s l}\right) - \theta_k \quad (2.9)$$

$$\theta_{shank} = \cos^{-1}\left(\frac{-\sin \theta_k}{l_s \cdot l}\right) \in (-1,1) \quad (2.10)$$

We know;

$$\alpha = \frac{\pi}{2} - \theta_k + \sin^{-1}\left(\frac{l_s \sin\theta_k}{l}\right) \quad (2.11)$$

During the onset of swing the knee torque generated; is given by;

$$\tau_k^i = \begin{cases} 0 & \dot{\alpha} > 0 \\ -k^i \alpha & \dot{\alpha} \leq 0 \end{cases} \quad (2.12)$$

Where $k^i$ refers to the flexion gain and $\dot{\alpha}$ refers to the rate of change of the leg angle,

$$\sin\left(\alpha - \frac{\pi}{2} + \theta_h\right) = \frac{l_s \sin\theta_k}{l} \quad (2.13)$$

$$l_s l \cos(\alpha + \theta_h) = \sin\theta_k \quad (2.14)$$

$$\cos^{-1}\left(\frac{\sin\theta_k}{l_s}\right) - \theta_k = \alpha \quad (2.15)$$

$$\tau_k^i = \begin{cases} 0 \\ -\frac{ki}{l}[\cos^{-1}(\sin\theta_k)]' - \dot{\theta}_k \end{cases} \quad (2.16)$$

As $l_s$ and $l$ are constants; where $l_s$ refers the length of the shank,

$$\frac{d}{dt}\left(\cos^{-1}(\sin\theta_k)\right) \quad (2.17)$$
$$= \frac{-1}{\sqrt{1-\sin^2\theta_k}} \cdot \cos\theta_k \cdot \dot{\theta}_k$$
$$= -\dot{\theta}_k$$

$$\tau_k^i = \begin{cases} 0 & \dot{\alpha} > 0 \\ \dfrac{2ki\dot{\theta}_k}{l_s l} & \dot{\alpha} \leq 0 \end{cases} \quad (2.18)$$

In cases where the passive knee flexion is inadequate, resulting in the foot swinging forward with a potential risk of scraping the ground, the control mechanism generates an active flexion torque at the knee joint. This torque is proportional to the rate of forward leg motion, denoted as $\alpha$.

The control mechanism responsible for regulating the movement of the knee employs a finite state machine that transitions between three distinct phases. In the initial phase, the knee is allowed to flex passively, meaning it moves without any active muscular effort, in response to the hip moments generated at the beginning of the swinging motion.

These set of constrained equation gives us some degree of interchangeability of the joint angles and thus help us in kinematic analysis and also helps in interpreting the torque during the swing phase with respect to shank angular velocity.

## IV. ELIMNATION OF JOINT ANGLE VARIABLE

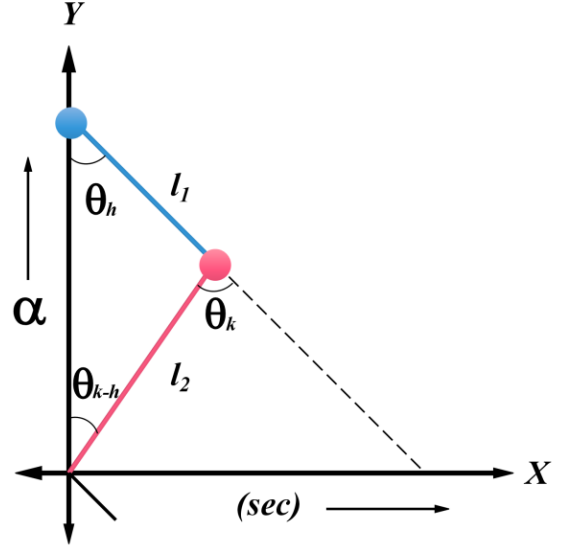

**Figure 4.** Computing Inverse Kinematic Parameters

Consider the instance at time $t = 1$;

At this point the relation between $\theta_1$ and $\theta_2$;

$$\sin\frac{(\theta_2 - \theta_1)}{l_1} = \frac{\sin\theta_1}{l_2} \quad (3.1)$$

At time $t = 2$;

$\theta_1$ and $\theta_2$ changes, set;

$$\theta_1 \to \theta_1\}_{At\ t=1} \quad (3.2)$$

$$\theta_2 \to \theta_2\}_{At\ t=2} \quad (3.3)$$

$$\frac{\sin(\theta_{1\}At\ t=1})}{l_2} = \frac{\sin(\theta_{2\}At\ t=1} - \theta_{1\}At\ t=1})}{l_1}$$
$$= \alpha \quad (3.4)$$

$\theta_2$ can be obtained continuously from the IMU as it changes with time thus $\theta_1$ at any instance can be calculated.

The '$l$' length of the vertical does not influence at any instant at $t = $ '$n$' seconds,

$$\frac{\sin(\theta_{1\}At\ t=n})}{l_2} = \frac{\sin(\theta_{2\}At\ t=n} - \theta_{1\}At\ t=n})}{l_1}$$
$$= \alpha \quad (3.5)$$

holds true.

Given the $\theta_h$ is the angular displacement of hip and $\theta'_h$ Angular velocity of the hip we can find the angular velocity of knee and finding. $\theta'_k$ will give the angular displacement of the knee.

Change in Torque position with respect to change in value position,

$$\tau_k = 0.0804\ddot{\theta}_k - 0.2553\ddot{\theta}_h(\theta_h + \theta_k) + 0.2553(\dot{\theta}_h)^2 \sin(\theta_h + \theta_k) + 0.032\ddot{\theta}_k + 4.57168\sin(\theta_k) \quad (3.6)$$

___

Applying our functions and grouping the like terms, we get,

$$\tau_{k_1} = \gamma(\dot{\theta}_k{}^2 + \tan \theta_k(0.9516) + \ddot{\theta}_k) \quad (3.7)$$

___

Similarly, the term $\ddot{\theta}_t(\theta_t + \theta_s)$ must be differentiated w.r.t to time

$$\tau_{k_2} = -0.2553\left\{\ddot{\theta}_k(1 + \gamma l_s l) + \ddot{\theta}_k \dot{\theta}_k l_s l \left[\frac{\sin 2\theta_k}{2}\left(\gamma - \frac{1}{\gamma}\right) - \tan \theta_k\right] - \frac{\gamma}{2} l_s l \sin 2\theta_k\right] \quad (3.8)$$

___

$$\tau_{k_3} = 0.2553(\theta_k + \alpha)(\dot{\theta}_k(1 + \gamma l_s)\sin(\theta_k + 2\alpha) + 1) \cdot \dot{\theta}_k\{(\theta_k + \alpha)\cos(\theta_k + 2\alpha)[1 + \gamma l_s l] + \dot{\theta}_k \gamma)] + (1 + \nu l_s l)\sin(\theta_k + 2\alpha) \quad (3.9)$$

___

$$\tau_{k_4} = 4.57168\,\dot{\theta}_k(\gamma \sin \theta_k) \quad (3.10)$$

___

| Parameters | Specifications |
|---|---|
| $\theta_k$ | Knee angle |
| $\dot{\theta}_k$ | Knee angular velocity |
| $\ddot{\theta}_k$ | Knee angular acceleration |
| $\theta_h$ | Hip angle |
| $\dot{\theta}_h$ | Hip angular velocity |
| $\ddot{\theta}_h$ | Hip angular acceleration |
| $\alpha$ | Leg angle |
| $\tau_k$ | Knee Torque |
| $l_S$ | Shank Length |
| $l$ | Length of the leg |

**Table 2.** Inverse Kinematic Parameters

## V. ESTIMATION OF SPATIAL PARAMETERS

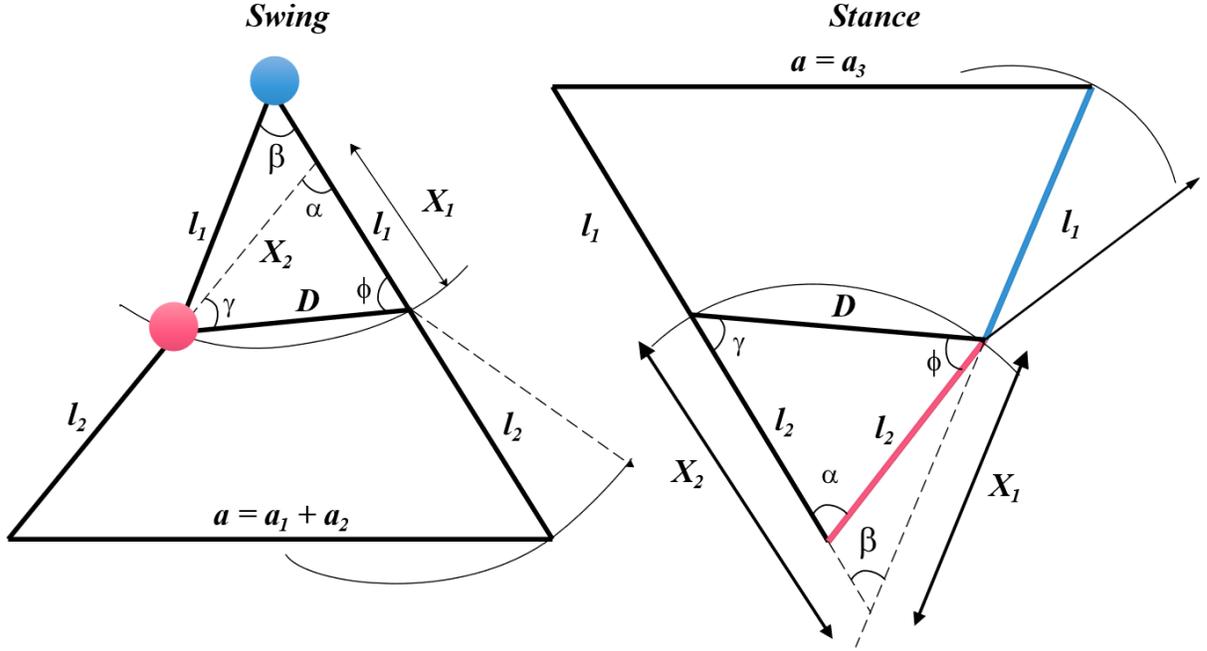

**Figure 5.** Body segments model of a gait cycle, enabling distance estimation based on angular rate signals.

In our research, we put forth a gait model that incorporates both the shank and thigh segments. This model takes into account the swing phase, which is represented as a double pendulum model, and the stance phase, which is represented as an inverse double pendulum model.

The combined distance traveled by the right shank and right thigh swing during a gait cycle is represented by the sum of a1 and a2. Furthermore, during this phase, the body moves forward by an additional distance d3 due to the rotational motion of the left shank and left thigh during the opposite stance phase. Assuming symmetry in step lengths, it can be inferred that the rotation experienced during the left stance phase is comparable to that of the right stance phase. Stride length (SL) incorporates these considerations:

$$G_c(n) = a_1 + a_2 + a_3 \tag{4.1}$$

By assigning the variables $\alpha$ and $\beta$ to represent the angular rotations of the right thigh and right shank, respectively, the distance a can be determined by summing up a1 and a2 through the application of trigonometric relationships. Specifically, during the swing phase of each gait cycle (referred to as k), the following computations are applicable.

$$a = a_1 + a_2 \tag{4.2}$$

$$a(x) = a_1(n) + a_2(n) \tag{4.3}$$

During the swing phase we have,

$$\frac{x_1(n)}{\sin \gamma(n)} = \frac{D}{\sin \alpha(n)} \tag{4.4}$$

$$\frac{x_2(n)}{\sin \phi(n)} = \frac{D}{\sin \alpha(n)} \tag{4.5}$$

$$2\phi_n + \beta_n = \pi \tag{4.6}$$

$$2\gamma_n + 2\alpha_n - \beta_n = \pi \tag{4.7}$$

The angles $\alpha$ and $\beta$ were obtained by performing integration on the angular rate rotations of the right thigh and right shank:

$$D = 2il_1 \cos(\beta/2) \tag{4.8}$$

$$\beta(n) = \int_{TO(n)}^{Hs(n)} R_S(t) dt \tag{4.9}$$

$$\alpha(n) = \int_{TO(n)}^{Hs(n)} R_t(t) dt \tag{4.10}$$

And during the stance phase we have,

$$\frac{x_1(n)}{\sin \gamma(n)} = \frac{D}{\sin \alpha(n)} \tag{4.11}$$

$$\frac{x_2(n)}{\sin \phi(n)} = \frac{D}{\sin \alpha(n)} \tag{4.12}$$

$$2\phi_n + \alpha_n = \pi \tag{4.13}$$

$$2\gamma_n + 2\beta_n - \alpha_n = \pi \tag{4.14}$$

$$D = 2il_2 \cos\left(\frac{\alpha}{2}\right) \tag{4.15}$$

$$\beta_n = \int_{HS(n)}^{TO(n)} R_t(t)dt \tag{4.16}$$

$$\alpha_n = \int_{HS(n)}^{TO(n)} R_s(t)dt \tag{4.17}$$

For swing phase we have,

___

$$a(x) = \sqrt{\left(L_{2X1}(x)\right)^2 + \left(L_{2X_2}(x)\right)^2 + (L_{2X1}(x))(L_{2X2}(x))\cos\alpha(x)} \tag{4.18}$$

___

$$A_{RSW} = h_L a_1 - A_{(2\alpha-\beta)} \tag{4.19}$$

___

$$A_{(2\alpha-\beta)} = \sqrt{S_{2\alpha-\beta}(S_{2\alpha-\beta} - l_1)(s_{2\alpha-\beta} - l_2)\left(s_{2\alpha-\beta} - \sqrt{l_1^2 + l_2^2 - l_1 l_2 \cos(2\alpha - \beta)}\right)} \tag{4.20}$$

___

$$40\%G_c(n) = \int_0^{l_2 \cos\gamma} \tan\gamma x\, dx + \int_{l_2 \cos\gamma}^{l_1(\beta+\phi)} \frac{l_2 \sin\gamma + l_1 \cos(\beta + \phi)}{l_2 \cos\gamma - l_1 \sin(\beta + \phi)}[x - l_2 \cos\gamma] + l_2 \sin\gamma\, dx \tag{4.21}$$
$$= A_{RSW}$$

___

For stance phase we have,

___

$$a(x) = a_3(x) = \sqrt{(L_{1X2}(n))^2 + (L_{1X1}(n))^2 + (L_{1X2}(n))(L_{1X1}(n))\cos\beta(x)} \tag{4.22}$$

___

$$60\%G_C(n) = \int_{a_1}^{\sin\beta_2(l_1+l_2)} \frac{l_1 \sin(\beta + \gamma)}{l_1 \cos(\beta + \phi) - \sin\frac{\alpha}{2}(l_1 + l_2)}(\text{x} - l_1\cos(\beta + \phi) + l_1 \sin(\beta + \gamma)\, dx + \frac{1}{2}a_1 h_L - A_{(2\alpha-\beta)} \tag{4.23}$$
$$= A_{RST}$$

| Parameters | Specifications |
|---|---|
| $G_c(n)$ | Stride length of one gait cycle |
| $A_{(2\alpha-\beta)}$ | Area of the isosceles triangle for gait length calculation |
| $A_{RSW}$ | Swing Area |
| $A_{RST}$ | Stance Area |
| $a(x)$ | Swing length |
| $L_{2X1}$ | Square of sum of lengths l2 and X1 |
| $L_{1X2}$ | Square of sum of lengths l1 and X2 |

**Table 3.** Spatial Parameters

## VI. DISCUSSION

In the beginning of this paper, we go through the dynamic model of lower limb prosthesis illustrated in figure 1. and proposed by Xiaodong Wang et.al.

The equation (1.6) is capable to produce the knee torque with the input of the dynamic parameters. table 1. lists the parameters involved in the dynamic modelling of the lower limb prosthesis.

Section 3 addresses a pure geometrical model of lower limb prosthesis as shown in figures 2. and 3. in which equation (2.10) gives a domain constrained relationship between the shank and the knee angle. This is fundamentally important as the state of art gait analysis done today requires the placement of gyroscope, accelerometers and IMU in the shank in any of the biomechanical planes. As the proposed equation clearly stands for one dimensional sagittal plane, the equation can be extended to all three planes with the concepts such as vectors and cross products. As these sensor modalities give the measure of the shank angle, musculoskeletal dynamic simulation software such as OpenSim is used to obtain the inverse kinematic and dynamic paramters by using a constrained model. However, creation of such models is extremely time-consuming process as it requires the proper calibration of the anatomical markers and must be done independently for different people to get an accurate measure. Thus, our equation (2.10) takes advantage of the fact that a human leg is a constrained double pendulum model and will continue to show a general trend of similar dynamics while walking and has predefined limits for the joint angle variables listed in table 2.

Section 4 of the paper deals with the mathematical approach in the estimation of Knee joint angle given the hip joint angle and vice-versa. The equation (3.5) holds true for level walking as the vertical remains intact throughout the gait cycle and uses the sine rule as depicted in figure 4. which proves that at any instance, provided we have a joint angle the second joint angle can be estimated and is independent of the change in length of the vertical, this concept could be further extended to finding the joint angular velocity and angular acceleration which could then be used in the dynamic equation of (1.6) to obtain the torque values. The torque equation is formed based on the work done by Kalyan et. Al (2017) [47].

The section 5 of the paper resolves the swing and stance phase model of a gait cycle into double pendulum and inverse double pendulum model respectively. This model shown in figure 5. makes use of the spatial parameters mentioned in table 3. to obtain the angular rate signals, determine the stride length and stride velocity by diving the length of the stride by gait cycle duration of leg. The equations (4.1) – (4.23) listed in this section also gives us the area covered by the legs in the sagittal plane in both the swing and stance phase in terms of

## VII. CONCLUSION

In this study, a dynamic model of the lower limb prosthetic system was established by designing an intelligent knee joint structure. The torque required for the knee prosthesis at different speeds are different. Using the anthropometric data and preliminary data required for determination of toques, to actuate the considered prosthetic leg, generalized torque equations have been derived with the help of mathematical modelling through Lagrangian equations using MATLAB/SIMULINK environment. The equations stand good for modified parameters as well, because the torques to be attained are entirely dependent on the input parameters and this will be helpful for the development of prosthetic leg.

The model can also be used to determine the motor torque required for a prosthetic leg to mimic the motions of a healthy human leg. It can be done by placing the input parameters like length of the links of the prosthetic leg, corresponding mass of each link, the joint angle and time relations. For specific lower limb amputation patients, data from healthy people with similar body size can be used as initial data for prosthetic wearers. After the prosthesis is worn, the angle information of the wearer's healthy leg and the knee parameters are recalculated to obtain an optimized opening degree.

In conclusion, it can be said that calculation of the dynamic torques has great importance to predict the torque demands of the motors, since the appropriate motor power can be estimated only after deriving these torque values.

.